\documentclass[letterpaper, 10 pt, conference]{IEEEtran}  

\IEEEoverridecommandlockouts                              

\usepackage{graphics} 
\usepackage{epsfig} 
\usepackage{mathptmx} 
\usepackage{times} 
\usepackage{amsmath} 
\usepackage{amssymb}  
\usepackage[T1]{fontenc}
\usepackage{amsfonts}
\usepackage{booktabs}
\usepackage{siunitx}
\usepackage{caption} 
\usepackage{booktabs}
\usepackage{graphicx}
\usepackage{subfig}
\usepackage{comment}

\title{\LARGE \bf
Classification of Spot-welded Joints in Laser Thermography Data using Convolutional Neural Networks
}

\author{Linh K{\"a}stner$^{1}$\thanks{$^{1}$Linh K{\"a}stner and Jens Lambrecht are with the Chair Industry Grade Networks and Clouds, Faculty of Electrical Engineering, and Computer Science,				
		Berlin Institute of Technology, Berlin, Germany
		{\tt\small linhdoan@tu-berlin.de}}, Samim Ahmadi,$^{2}$\thanks{$^{2}$Samim Ahmadi, Florian Jonietz and Mathias Ziegler are with the Bundesanstalt f{\"u}r Materialforschung, Berlin, Germany
		{\tt\small samim.ahmadi@bam.de}} Florian Jonietz$^{2}$, Mathias Ziegler$^{2}$,\\ Peter Jung$^{3}$, \textit{IEEE member}\thanks{$^{3}$Peter Jung and Giuseppe Caire are with the Chair Communication and Information Theory, Berlin Institute of Technology, Berlin, Germany}, Giuseppe Caire$^{3}$, \textit{IEEE fellow}, and Jens Lambrecht$^{1}$
}

\begin{document}

\maketitle
\thispagestyle{empty}
\pagestyle{empty}


\begin{abstract}

Spot welding is a crucial process step in various industries. However, classification of spot welding quality is still a tedious process due to the complexity and sensitivity of the test material, which drain conventional approaches to its limits. In this paper, we propose an approach for quality inspection of spot weldings
using images from laser thermography data. We propose data preparation approaches based on the underlying physics of spot welded joints, heated with pulsed laser thermography by analyzing the intensity over time and derive dedicated data filters to generate training datasets. Subsequently, we utilize
convolutional neural networks to classify weld quality and
compare the performance of different models against each other. We
achieve competitive results in terms of classifying the different
welding quality classes compared to traditional approaches, reaching an accuracy of more than 95 percent. Finally, we explore the effect of different augmentation methods.

	
\end{abstract}

\section{INTRODUCTION}
Spot welding plays a major role in joining technologies, especially in the automotive industry. Traditional methods to assure the quality of spot welded joints include random and periodic destructive tests like torsion testing or manual destructive testing, where the specimen has to be cut in half to be investigated. 
These methods are tedious and destroy the sample. Non-destructive testing methods (NDT) reduce the costs of quality assurance and imply an optimization of the method of spot welding, since every joint could be checked, and therefore the number of spot welded joints could be reduced. 
Among popular NDT methods for quality inspection of welded material are ultrasonic testing, X-Ray tomography \cite{saravanan2014non}, acoustic emission testing and laser thermography.
X-Ray has been considered as reliable approach to assess the welding quality. Kar et al. \cite{kar2018x} used X-Ray tomography to study the porosity of welded joints and asses the quality. Patil et al. \cite{patil2016investigation} investigated weld defects using X-Ray radiography and found that the X-Ray method could reveal more defects compared to a visual inspection.
While X-Ray approaches are a commonly used NDT method, the necessary radiation protection is a major limitation, thus it cannot be easily applied for in-situ inspection. In addition, X-Ray computer tomography is expensive compared to other NDT methods such as ultrasound or thermography. Furthermore, the wave's penetration degree is limited, especially with multi-layered material thus could not be applied to detect small defects as observed by Duchene et al. \cite{duchene2018review}.
As an alternative, ultrasonic approaches are being increasingly considered.
Yu et al. \cite{yu2019detection} proposed an approach which employed high order ultrasonic waves to detect damages in welded joints and thus, could enhance the detection sensitivity to detect small weld flaws.
Tabatabaeipour et al. \cite{tabatabaeipour2016non} proposed an immersion ultrasonic testing method by observing the backscattered energy C-Scan images.
Papanikolaou et al. \cite{papanikolaou2020non} used ultrasonic testing as NDT method to inspect various parameters such as the chemical compositions or mechanical properties of the specimen to determine the weariness of specimen. The researchers conclude enhanced results using ultrasound testing, compared to visual testing and liquid penetration testing.
Acoustic approaches on the other hand, utilizes ultrasonic waves at a much higher frequency and have been employed by a variety of work. 
Shrama et al. \cite{sshrama2019use} applied acoustic emission to inspect welded joints for damages. They conducted a variety of tests and conclude an enhancement in understanding of damage mechanism for early maintenance. Kubit et al. \cite{kubit2019analysis} utilized acoustic microscopy to evaluate the joint quality. Despite its increased sensitivity, the setup and operation is very complex.
Active thermography, on the other hand, emanates in recent times as a method, which allows contactless, fast and reliable testing, at cheaper operation costs than e.g. computer tomography. The feasibility of spot weld inspection based on thermography was theoretically examined in \cite{siemer2010einsatz}. In \cite{myrach2017calibration}, the researchers could already show that thermography is a robust alternative and can be calibrated using X-Ray methods. A non-destructive testing approach based on laser thermography was proposed by Jonietz et al. \cite{jonietz2016examination}, where the researchers could detect important metrics of quality like the welding diameter by applying active thermography in transmission and reflection. However, the quality of the spot welded joints could not be assessed in detail. \\
Convolutional neural networks (CNNs) have achieved remarkable results in computer vision for tasks such as anomaly detection and classification, thus gaining immense popularity in NDT research in recent years. Cruz et al. \cite{cruz2017efficient} used CNNs to detect defects in ultrasound testing. Works by \cite{suyama2019deep}, \cite{wang2019welding} and \cite{faghih2016deep} use CNNs to detect welding defects within X-Ray images and show performance enhancements. 
For instance, Wang et al. \cite{wang2019welding} used a RetinaNet-based CNN architecture to detect and classify three different types of defects inside X-Ray images.
Zhang et al. \cite{zhang2019weld} presented a weld defect detection on X-Ray images based on CNNs. The researchers achieve satisfying results in detecting features relevant for quality assessment. However, since the X-Ray approach is based on the transmission of radiation through the spot-welded joint, it is only possible with access from both sides. 
Janssens et al. \cite{janssens2017deep} explored the usage of a deep neural network on infrared thermal images to monitor machine health by detecting fault conditions from moving machine components. The researchers conclude a significant performance boost when applying CNNs and that relevant regions could be identified and visualized to detect potential failures. Nasiri et al. \cite{nasiri2019intelligent} used CNNs to detect six conditions in thermal images of cooling tubes.
Similar to our work, Yang et al. \cite{yang2019infrared} used a Faster-RCNN-based architecture to visualize defects inside metal plates inducted with heat. They analyse the heat distribution and propose an improved Faster-RCNN architecture to visualize and detect the cracks. 
Dung et al. \cite{dung2019vision} explored the effect of CNNs on welded joints on gusset plates and conclude its feasibility when using transfer learning and data augmentation. In this paper, we will utilize both methods as well but will specify data preparation methods specifically for thermography data. Therefore, we consider the underlying physics such as the heat distribution and the temporal component of thermal images, which provide more information about the specimen. In addition, our data acquisition approach is contactless and, very importantly, requires access to the weld from one side only, making it proficient for in-situ quality inspection.
For improved feature visualization, we apply preprocessing steps presented in \cite{jonietz2016examination}.

The main contributions of this work are the following:
\begin{itemize}
    \item Proposal of CNN-based welding quality assessment method to classify welding quality from thermal images that are not distinguishable by human vision inspection.
    \item Proposal of methods to generate a feasible training dataset from thermal images by analyzing the underlying physics and generating filters accordingly.
	\item Evaluation of different data augmentation methods and their effect on thermal datasets.
	\item Performance evaluation of three State-of-the-Art neural network architectures.
\end{itemize}

The paper is structured as follows. Sec. II begins with the theoretical foundations utilized in our approach. The methodology including the overall concept and the implementation of each module, is presented in Sec III. Sec. IV presents the results and discussion. Finally, Sec. V will give a conclusion and outlook.

\section{Theoretical foundations}
The data analyzed in this contribution has been acquired using laser thermography. The theoretical backgrounds are presented in this chapter. 

\begin{figure}
    \centering
    \includegraphics[scale = 0.7]{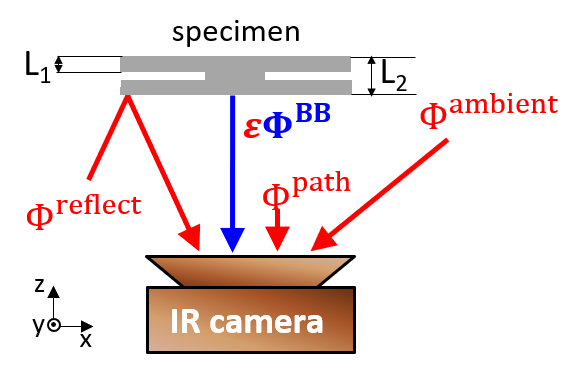}
    \caption{Sketch for theoretical understanding of IR radiant flux in our experiments. The IR camera receives different IR radiation components (red arrows), whereas the direct component from the specimen is given as a blue arrow. The schematic of the cross section through a welding joint is given on top in gray color.}
    \label{fig:theory}
\end{figure}

\subsection{Description of the thermal radiation components and emissivity corrections} \label{sec:thermalradiation}
Fig. \ref{fig:theory} illustrates our setup for a theoretical understanding of the IR  radiant flux used in our experiments.
The radiant flux $\Phi$ (SI unit: Watt) is a common quantity to describe the intensity level of the IR radiation. Fig. \ref{fig:theory} shows the IR radiation components as detected by the IR camera in the measurement environment: Direct radiation from the ambient environment ($\Phi^{\text{ambient}}$), environmental radiation reflected from the surface of our investigated specimen ($\Phi^{\text{reflect}}$), as well as radiation from the measurement path between specimen and our measurement device ($\Phi^{\text{path}}$), which is caused by the atmospheric absorbers (e.g., air, humidity, CO$_2$). All these disturbing quantities (summed up in the following as $\Phi^{\text{env}}$) are detrimental for our measurement since we are interested in measuring the radiant flux of the specimen $\Phi^{\text{specimen}}$. In addition, we do not know the exact emissivity $\epsilon$ of our specimen that indicates how much radiation it emits compared to an ideal heat radiator, i.e. a black body (BB).
These conditions, lead to the following total radiant flux $\Phi^{\text{tot}}$ during our measurements for every pixel:
\begin{align}
\label{eq:total_heat}
    \Phi^{\text{tot}} &= \Phi^{\text{env}} + \Phi^{\text{specimen}} \\
               &= \Phi^{\text{env}} + \epsilon \Phi^{\text{BB}}. \nonumber 
\end{align}
The emissivity $\epsilon$ is a unit-less scalar with $\epsilon \in [0,1]$.
According to Stefan-Boltzmann law, the radiant flux $\Phi$ depends on the temperature ($\Phi \propto T^4$). During a thermographic measurement we can rewrite $\Phi^{\text{BB}}$ to $\Phi^{\text{BB}}_{T(t)}$ and before the measurement, we can write $\Phi^{\text{BB}}_{T_0}$ with $T_0$ standing for the room temperature and $T(t)$ considering the temporal heating given by laser illumination ($t>0$). 
Assuming constant environmental conditions and temperature-independent optical quantities of the specimen, $\Phi^{\text{env}}_{T_0} \approx \Phi^{\text{env}}_{T(t)}$ and $\epsilon$ remain the same during the experiment. The environmental disturbances $\Phi^{\text{env}}$ could be therefore removed if we consider the radiant flux difference $\Phi^{\text{tot}}_{T(t)} -\Phi^{\text{tot}}_{T_0}$. Further, the unknown emissivity $\epsilon$ can be removed by considering a normalized radiant flux difference$(\Phi^{\text{tot}}_{T(t)} -\Phi^{\text{tot}}_{T_0})/(\Phi^{\text{tot}}_{T(t_{\text{norm}})} -\Phi^{\text{tot}}_{T_0})$ where $t_{\text{norm}}$ refers to another time after the sample is cooled down to a temperature $T(t_\text{norm})>T_0$. For more detailed explanations we refer to \cite{jonietz2016examination}. In this contribution we utilized this method to generate a noise-free dataset without any uncertainties due to the emissivity. 
Please note that in this approach we have to calculate with the temperature dependent radiant flux (as measured with the IR camera) and not with the temperature (calculated inside the IR camera based on a previous calibration) itself. Moreover, using Stefan-Boltzmann law is an approximation, since the IR camera is sensitive in a restricted spectral range only.

\subsection{Data description} \label{sec:datadescription}
\begin{figure}[!h]
	\small{(a)}
	\includegraphics[width=0.45 \textwidth]{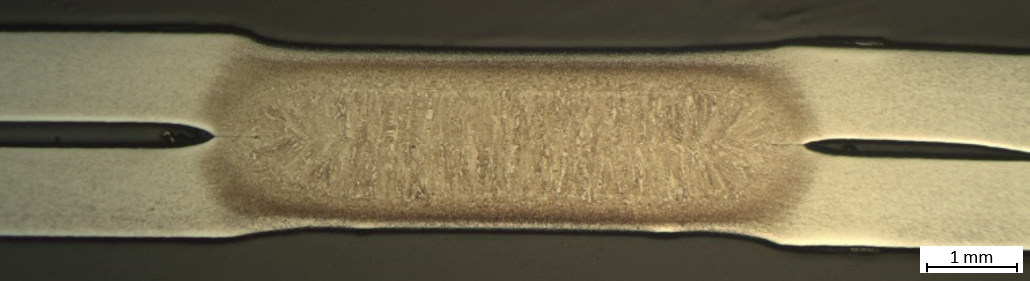}
	\small{(b)}
	\includegraphics[width=0.45 \textwidth]{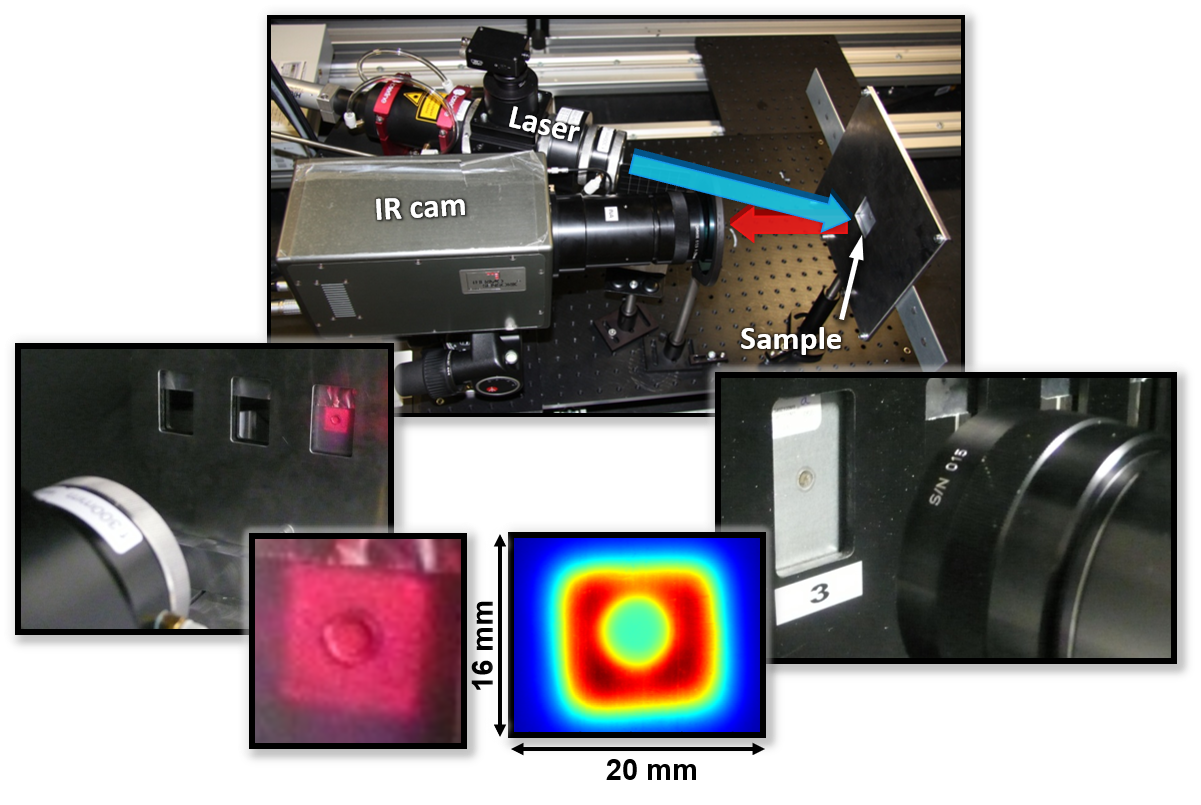}
	\caption{ (a) Metallography of one of our specimens after applying resistance spot-welding. (b) Data acquisition setup. The specimen, consisting of two welded metal sheets, is heated up with active laser thermography. The heat distribution over time is measured with the IR camera. The diameter of the spot-welded joints is around 4-8 mm.}
	\label{experiment}
\end{figure}

In our experiment, we perform pulsed thermography using a rectangular shaped homogeneous laser illumination over the whole area of interest. Therefore, we can calculate the 2D solution (referring to the two spatial dimensions $x$ and $y$, see Fig. \ref{fig:theory}) for the homogeneous heat diffusion equation for a 2D heating source in reflection configuration ($z=0$) by \cite{PSF}:
\begin{align} \label{eq:HDE}
    T(x,y,t) = T_0 + \frac{Q/A}{\rho c_p \sqrt{\pi \alpha t}} e^{-\frac{(x+y)^2}{4\alpha t}} \bigg(1+ 2\sum_{n=1}^{\infty} R^n e^{-\frac{(nL)^2}{\alpha t}}\bigg)
\end{align}
whereby $Q$ describes the absorbed radiation energy from the laser, $A$ the illuminated area, $t$ the time, $\rho$ the material density, $c_p$ the specific heat capacity of the material, $\alpha$ the diffusivity, $R$ the thermal reflectivity (material to air), $n$ the number of reflections of the so-called thermal wave and $L$ the thickness of the plate. The given temperature evolution refers to an ideal sheet infinitely extended in the plane and an infinitely short heating impulse. For the actual specimen and experimental setup, we can only get a first impression on how the temperature evolves and concentrate on the transient signal contrast in the dataset caused by the geometry of the specimen. Fig. \ref{fig:theory} shows that the value for $L$ differs since the specimen consists of two steel sheets welded together by a spot-welded joint. This means that, according to eq. \eqref{eq:HDE}, the solution for the heat diffusion equation in the area of the spot-welded joint works with $L=L_2$ whereas the region outside the spot-welded joint works with $L=L_1$. Fig. \ref{experiment} (b) shows also the main difference of the heat flow visually (red - high temperature, blue - low temperature). Since we are measuring in reflection configuration (IR camera and laser on the same side of the specimen), we observe a hot rim outside the spot-welded joint region since the heat is accumulated. On the other hand, we observe a cold spot in the middle since the heat diffuses through the spot-welded joint towards the other steel sheet. Therefore, a good connection should serve for an evident contrast between the region inside and outside the spot-welded joint. 

\begin{figure*}[!h]
  \centering
  \subfloat[][]{\includegraphics[width=3.06in]{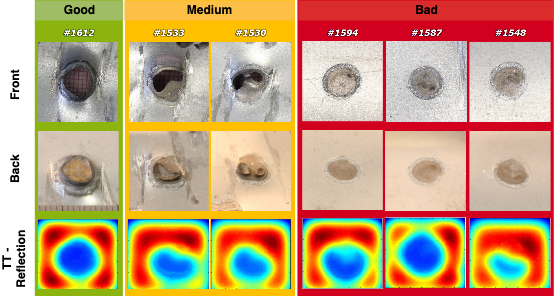}}%
  \quad
  \subfloat[][]{\includegraphics[width=3.6in]{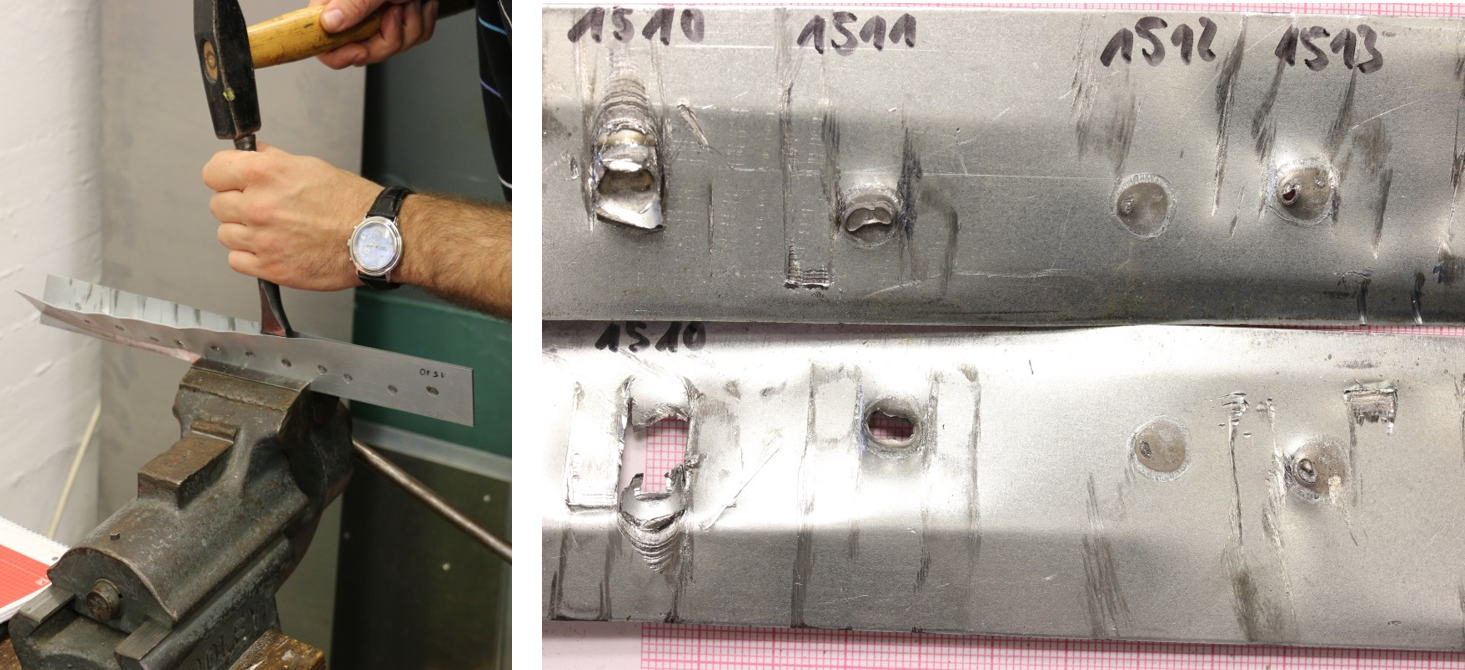}}%
  \caption{(a) Classification of welding quality. Specimens for the three different quality classes and their respective normalized thermal image are shown. It is evident that the classes appear similar. For instance, it is hard to classify between image 1612 and 1587. (b) Destructive testing as reference method to assure the quality of the spot-welded joints using hammer and chisel.}%
  \label{groundtruth}
\end{figure*}
In the following, we are working with intensities $I_{n,p} \in \mathbb{R}^{N_t}$, where $N_t$ designates the number of time stamps which is equal to the number of measured thermal images (thermograms) in a thermal film sequence. These intensities refer to the radiant flux of the thermal radiation as measured by the InSb-based detector of the IR camera and converted to digits using an analog-to-digital converter. Thus, in this work, the intensity values in a thermal image are given by digits pixelwise representing the measured radiant flux in $x$ ($n \in \{1, \dots, N_x\}$) and $y$ ($p \in \{1, \dots, N_y\}$).

\section{Methodology}
After describing the underlying physics and theoretical foundations of our data, in this section, we will present the methods that we used for our proposed quality assessment use case.
\subsection{Experimental setup and data acquisition}
Our dataset was collected from specimens that were made using an electric welding system, see Fig. \ref{experiment}. These specimens consist of two resistance spot-welded hot-dip galvanized micro alloyed steel sheets HX340LAD \cite{en1} (zinc layer is approximately 7.5\,$\mu\text{m}$ on each side), respectively, which are typically used in automotive industry and have a thickness of $1$\,mm. The resistance spot-welding has been performed using a welding current of $7.5$\,kA, a pressure of $3.5$\,kN, and a welding time of $240$\,ms using an electrical spot welding machine. According to the procedure for the determination of the electrode life \cite{iso2}, more than 1600 spot weldings have been performed. After approximately 1000 welds, the electrode life has been reached and started to produce unreliable spot-welded joints. We tested 115 welds using thermography starting from weld no. 1510. As reference, we applied destructive chisel testing according to Ref. \cite{iso200510447}. The setup for data acquisition is illustrated in Fig. \ref{experiment} (a). We used active laser thermography for all tested specimens and captured 250 frames over time, which results in a film for every test object visualizing the spatial heat distribution for each time step. The laser radiation was switched on for a duration of one second at $500$\,W, illuminating a square-shaped area of $19\times19\,\text{mm}^2$. The thermal images were measured with an IR camera (InSb detector, sensitive between $3.7-5.3$\,$\mu$m, frame rate: $40$\,Hz, spatial resolution varied between $62.5$ and $133$\,$\mu$m/pixel). The utilized fiber-coupled laser emits in the near infrared range ($940$\,nm) and is therefore not interfering with the detector range of the IR camera. The laser heats up the specimen with a spot-welded joint. As can be observed in Fig. \ref{experiment}, the challenge of our thermal dataset is the similarity of the raw infrared data for different quality classes, which is not distinguishable by human visual inspection. For instance, it is hard to classify between image 1612 and 1587 or 1533 and 1548, despite their different classes. The features specifying each class are not evident, which causes common feature extractors like CNNs to struggle with. One that account, we explore ways to generate feasible datasets out of utilizing the underlying physics of the laser thermography process described in the previous chapter.

\begin{figure*}[!h]
	\centering
	\includegraphics[width=6.6in]{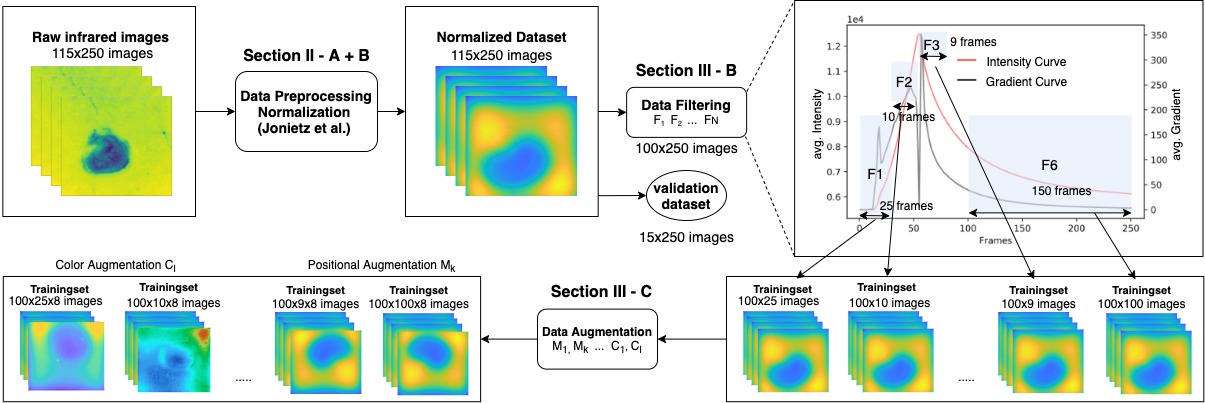}
	\caption{Data engineering workflow with exemplary filters illustrated in the data filtering section.}
	\label{wf2}
\end{figure*}

\subsection{Data filtering}
One of the aspects of this work is to explore how to process the normalized intensity data described in the previous section (see section \ref{sec:datadescription}) to provide reliable predictions using CNNs.
Therefore, we study different filters and their effect on the performance of the CNN.
We only extract certain images defining a filtered set $S^i_{\text{filt}} \subset S = \{1, \dots, N_{t}\}$ with the cardinality $|S^i_{\text{filt}}| = N_{\text{filt}}< N_t$. $\Phi$ can be replaced by $I_{n,p}$, referring to eq. \eqref{eq:total_heat}, and further it can be described by a 1D array with $I_{n,p}\in \mathbb{R}^{N_t}$ and $I = \begin{bmatrix}
I_{1,1} & \dots & I_{1,N_y} \\ 
\dots & \dots & \dots \\ 
I_{N_x,1} & \dots & I_{N_x,N_y} \\ 
\end{bmatrix} \in \mathbb{R}^{N_x \times N_y \times N_t}$, so that we can describe the filtered data by:
\begin{align}
F_i = I_{n,p}^{\text{norm}}[S^i_{\text{filt}}]
\end{align}
whereby $F_i$ represents a subset of the whole measured dataset with the specified intensity values defined in Table \ref{filterdatasets} so that $F_i$ denotes a filtered dataset. $I_{n,p}^{\text{norm}}$ stands for the normalized intensity difference as similarly described for the radiant flux in section \ref{sec:thermalradiation}.
Thus, the filters are defined based on intensity values of the films. 
These filters can lead to positive effects as we are investigating a dynamic temperature behaviour over time. Extracting only frames with significant changes in their amplitude, e.g. while heating or beginning of cooling phase, allows for more evident features within the datasets. The intensity is calculated by using the average value of all pixels in the image. Fig. \ref{wf2} (upper right corner) illustrates the intensity and gradient values referring to the temperature-time diagram as well as marked areas of filters and resulting datasets. For the generation of our final results, we use a combination of different filtered sets which yields
$(F_i, \, F_j, \, \dots,\, F_n)$
whereby $i$, $j$, $n \in \{1,\, \dots,\, 12\}$ and $F_i$, $F_j$, $F_n$ designate different filtered datasets according to Table \ref{filterdatasets}. 
In total, we define 12 different filtered datasets for the whole film each representing a different status of heating to investigate the effects of certain areas of the intensity curve on the performance of the CNN. The image counts of each dataset before and after augmentation are listed in Table \ref{filterdatasets} as (before || after). The applied augmentation methods are described in the next section.

\begin{table}[!h]\centering
	\caption{Image count for datasets}
	\begin{tabular}{@{}rrrrcrrrcrrr@{}}\toprule& \multicolumn{3}{c}{Filter Description} 
	 \phantom{abc}\\\cmidrule{1-5}\textbf{Filters} & $Frames$ & $Intensity$ & $Image Count$ & $Description$ \\\midrule 		
		$F_{1}$ & 1-25 & 5k-7k& 2.5k || 20k & No Heating	\\
		$F_{2}$ &  35-45 &9k-11k & 1k || 8k  &Intensity Peak 2 \\
		$F_{3}$ &  51-60& 12.5k-14.4k & 900 || 4560& Maximum  Intensity\\
		$F_{4}$ &  61-75& 9.7k-11k& 1.4k|| 11.2k & After-Maximum\\
		$F_{5}$ &  76-100& 8k-9.1k & 2.4k || 19.2k & Cool Down\\
		$F_{6}$ &  101-135& 7k-8k & 3.4k || 27.2k& Cool Down\\
		$F_{7}$ &  136-170& 6.6-7k & 3.4k || 27.2k & Cool Down\\
		$F_{8}$ &  171-210& 6.2-6.5 & 3.9k || 31.2k & Cool Down\\
	    $F_{9}$ &  211-250& 6.05-6.1k & 3.9k || 31.2k & End\\
		$F_{10}$ &  20-75 & 8k-14.4k & 5k || 40k & Peaks Combined\\
		$F_{11}$ &  1-250& 0.4-14k & 10k || 80k & Cool Down\\
		$F_{12}$ &  101-250& 6.2k-8k & 7.5k || 60k & Cool Down \\

		\bottomrule
	\end{tabular}

	\label{filterdatasets}
\end{table}

\subsection{Data augmentation}
It is well-known in data science that data augmentation techniques such as scaling, rotation and flipping yields a better data basis for the application of CNNs. We first filter the data to obtain a set $F\in\mathbb{R}^{N_\text{filt}}$ and then augment $F$ yielding a new set
$F^{\text{aug}} \in \mathbb{R}^{N_{\text{aug}}}$:
\begin{align} \label{augment}
    F^{\text{\,aug}} = [M_1(F),\dots,M_k(F),C_1(F),\dots,C_l(F) ]
\end{align}
where $M_1,\,\dots,\,M_k:\mathbb{R}^{N_\text{filt}}\rightarrow\mathbb{R}^{N_{\text{aug},k}}$ are coordinate transformations and $C_1,\,\dots,\,C_l:\mathbb{R}^{N_\text{filt}}\rightarrow\mathbb{R}^{N_{\text{aug},l}}$ represent color transformations which change the intensity values of a pixel within a film. More specifically, in this work we use $k = l = 3$ by employing following data augmentation techniques:
\begin{equation}
        \begin{cases}
    M_{1/2/3}, & \mbox{for horizontal\,(1)/\,vertical flip\,(2)/\,rotation\,(3)} \\
    C_{1/2/3}, & \mbox{for PCA-Color/\,hue\&\,saturation/\,illumination} \\
    \end{cases} 
\end{equation}
\subsection{Data Labeling}
Three classes are to be considered for classification: \textit{good, bad} and \textit{medium}. Fig. \ref{groundtruth} (a) describes the labeling benchmark on which the data labeling is based. This benchmark was created with destructive testing using the standardized chisel testing \cite{iso200510447} by destroying the specimen and inspecting the welding quality visually by a human expert (s. Fig. \ref{groundtruth} (b)). As a result, each image in our dataset contains a label stating whether it has good (standard spot weld diameter), bad (stick weld, i.e. no or only minimum actual spot weld) or medium (undersized weld nugget leading to a weak mechanical joint) welding quality.

\subsection{Neural Network Design}
\begin{figure}[!h]
	\centering
	\includegraphics[width=3.4in]{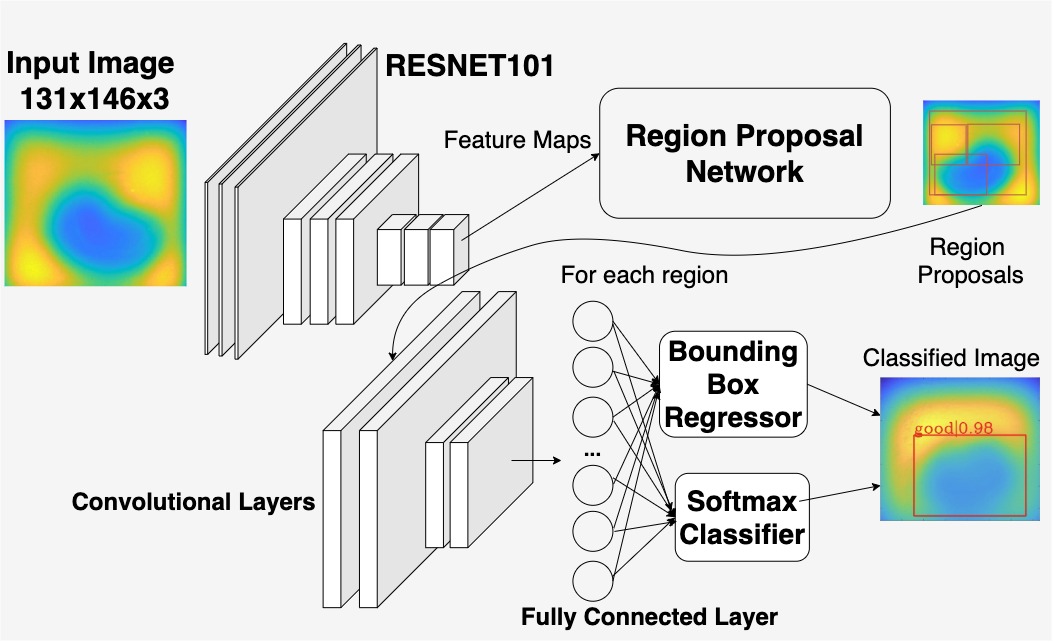}
	\caption{Architecture of our neural network}
	\label{nn}
\end{figure}

The data engineering steps previously discussed enable us to generate a feasible dataset with evident features for a convolutional neural network to robustly assess the welding quality. 
An important aspect of our data is that the frames starting approximately from frame 100 (after the cooling down phase) immediately become similar to each other and are not distinguishable. Since the areas of interest only contain 7-12 frames, a long short-term memory (LSTM) based approach which considers the temporal dependency would not deliver the desired results. The incorporation of recurrent neural networks was not considered because of the dominance of similar-looking frames which compromised nearly 80 percent of the film. Furthermore, the dynamics of our dataset is too low with only small changes visible between the frames. However, our observations also find that especially for the relevant areas like the maximum intensity area, the features will get evident for each class. Based on these considerations, a Faster-RCNN based 2D convolutional network is employed which will analyze one dedicated frame of the film, to make the prediction. Our architecture is based on the original Faster-RCNN \cite{ren2015faster} with modified input to match our thermal data and a ResNet101 as a backbone network. The architecture is visualized in Fig. \ref{nn}.
The input image has three channels and is of size $131\times146\times3$. As a backbone network, the ResNet-101 is employed. After passing the backbone network, feature maps are generated, which are passed through the region proposal network. Subsequently, for each region proposal, a bounding box regressor and a softmax classifier is applied to detect and locate the defects. The architecture is illustrated in Fig. \ref{nn}.

\section{Results and Discussion} 

\subsection{Filters Evaluation}
\begin{figure}[!h]
	\centering
	\includegraphics[width=3.3in]{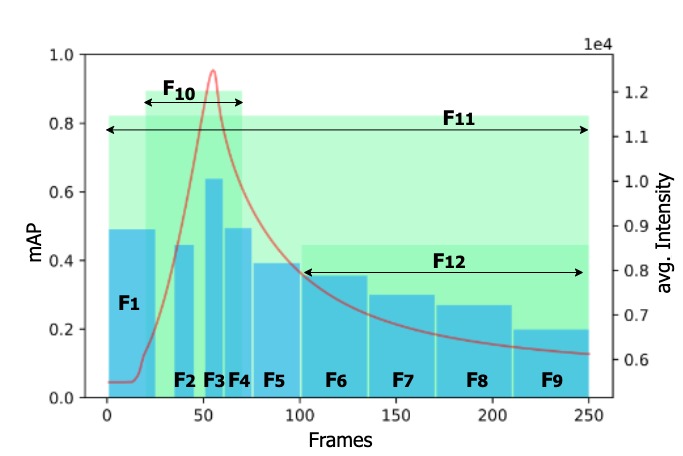}
	\caption{Accuracies of different intensity areas. The red curve is the average intensity curve of all films on which the filters are defined. The bars represent the according models' accuracies. Depending on the test dataset, the accuracy vary due to the more evident features of specific areas.}
	\label{filter2}
\end{figure}

\begin{figure}[!h]
	\centering
	\includegraphics[width=3.0in]{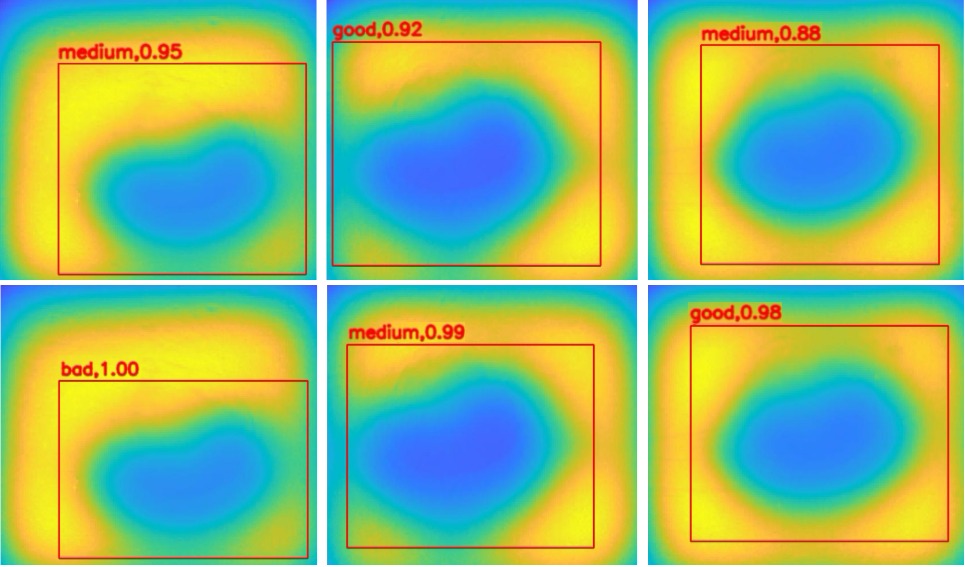}
	\caption{Performance comparison between different models. The upper row of the film results showcases the detection for three different films of model $F_{11}$, which were classified falsely while the lower row shows the detections of the same frames of model $F_{10}$ which were classified correctly}
	\label{comparisson}
\end{figure}

We trained the CNN with different datasets generated by applying the filters introduced in section III. Furthermore, the positional data augmentation techniques $M_1, M_2, M_3$ defined in Sec. III were applied: The images were horizontally and vertically flipped and rotated with a random value between -90 and 90 degrees. Since the heat diffusion is pointsymmetric, these positional changes will not affect the original information of the frame. 
Fig. \ref{filter2} illustrates the accuracies for the different datasets, each representing an intensity area.
We used the mean Average Precision (mAP) as evaluation metric, which indicates the classification probability of a correct result for a bounding box overlap of 50 percent to the groundtruth label (intersection over union = 0.5). The average of the accuracies for all three classes were calculated.
The highest accuracy is observed when using frames at large intensity values to train. However, using the same frames within smaller chunks of data, results in a significantly decreased performance. On that account, the effect of a combined dataset is explored by combining multiple filters as well as using the whole dataset for training. An evident accuracy boost can be observed while using the filtered dataset $F_{10}$ with images from frames of maximum intensity. However, it is noticeable that the smaller datasets from within the same area of intensity ($F_2,F_3,F_4$) result in significantly worse accuracies compared to the combined dataset ($F_{10}$). This observation is also evident when combining the datasets of frames 100 to 250, when the specimen is in its cool down stage. The results, albeit being already bad with only 30-40 percent accuracy, gain a small boost to 42 percent accuracy when being combined. However, since the specimen state at the end of a film is already cooled down completely, the visual differences between frames perish. Thus, the results are in line with our theoretical statements from chapter II. Therefore, they should not be considered when training the CNN as the similarity of the training data affects the performance of the CNN in a negative way.
Overall, we could improve accuracy by 6 percent when specifying filters which consider frames from the maximum intensity area of the film ($F_{10}$ compared to $F_{11}$). Interestingly, using the whole film does not decrease the accuracy significantly. As expected, areas at the end of the film will result in imprecise results with an accuracy of 40 percent. 
Fig. \ref{comparisson} showcases the detections of the two best achieving models resulting from dataset $F_{10}$ and $F_{11}$. While most test films could be classified correctly, there are some cases, in which the $F_{11}$ model gives a wrong prediction while the $F_{10}$ model could classify it correctly.

\subsection{Data Augmentation Evaluation}
\begin{figure}[!h]
	\centering
	\includegraphics[width=3.3in]{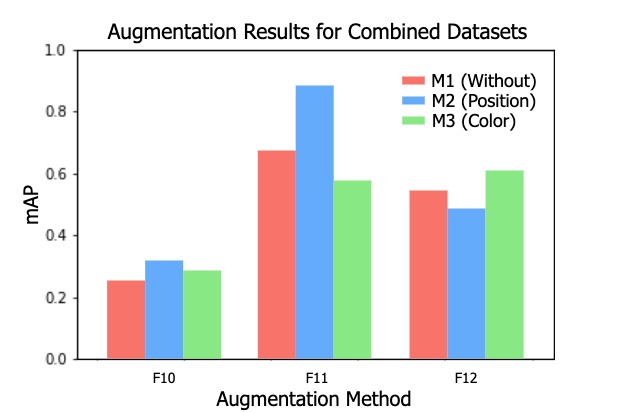}
	\caption{Impact of different augmentation methods on combined datasets ($F_{10},F_{11},F_{12}$)}
	\label{ar}
\end{figure}

To evaluate the impact of data evaluation methods, we applied different positional as well as color augmentations as defined in Sec. III. The results are depicted in Fig. \ref{ar}. For the color augmentation, the brightness and contrast saturation and PCA-Color was changed with a random value each. 

The accuracy could be improved when using the positional augmentations compared to the dataset without augmentation techniques applied.  This is more evident in the datasets $F_{10}$ and $F_{11}$. Since the heat diffusion is pointsymmetric, positional changes like rotation or flipping will not falsify the information inside the images. As expected, color augmentations affect the accuracy in a negative way for the datasets $F_{10}$ and $F_{11}$. Notably, the effect was not as evident as assumed. It is most evident in the area of maximum intensity, where the color augmentation decreased the accuracy. The area at the beginning of the cooling phase experiences a performance increase even when using color augmentation. This indicates a potential boost when using color augmentation due to the similar intensity values at later stages of the cooling down stage. The observed decrease in accuracy at stages where the intensity value is high, is due to the more evident spatial differences between frames, which a color augmentation would only disturb.

\subsection{Comparison with other CNN architectures}
\begin{figure}[!h]
	\centering
	\includegraphics[width=3.4in]{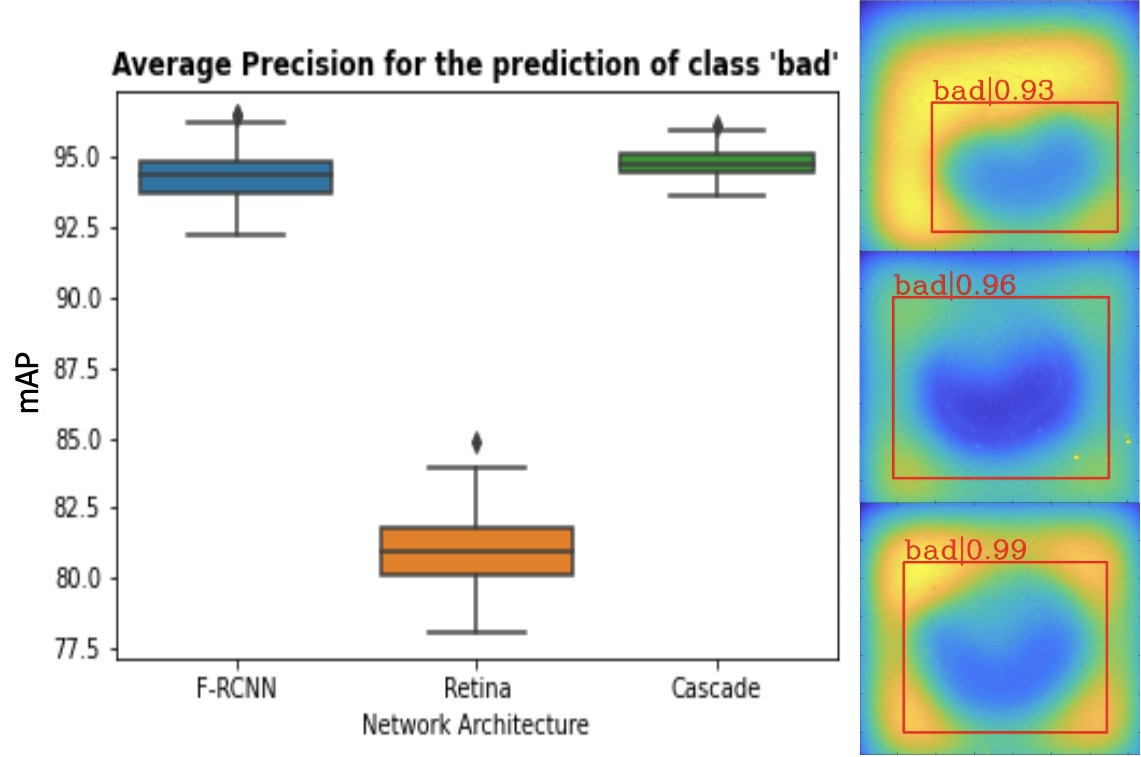}
	\caption{Detection results for class 'good'}
	\label{res1}
\end{figure}

\begin{figure}[!h]
	\centering
	\includegraphics[width=3.4in]{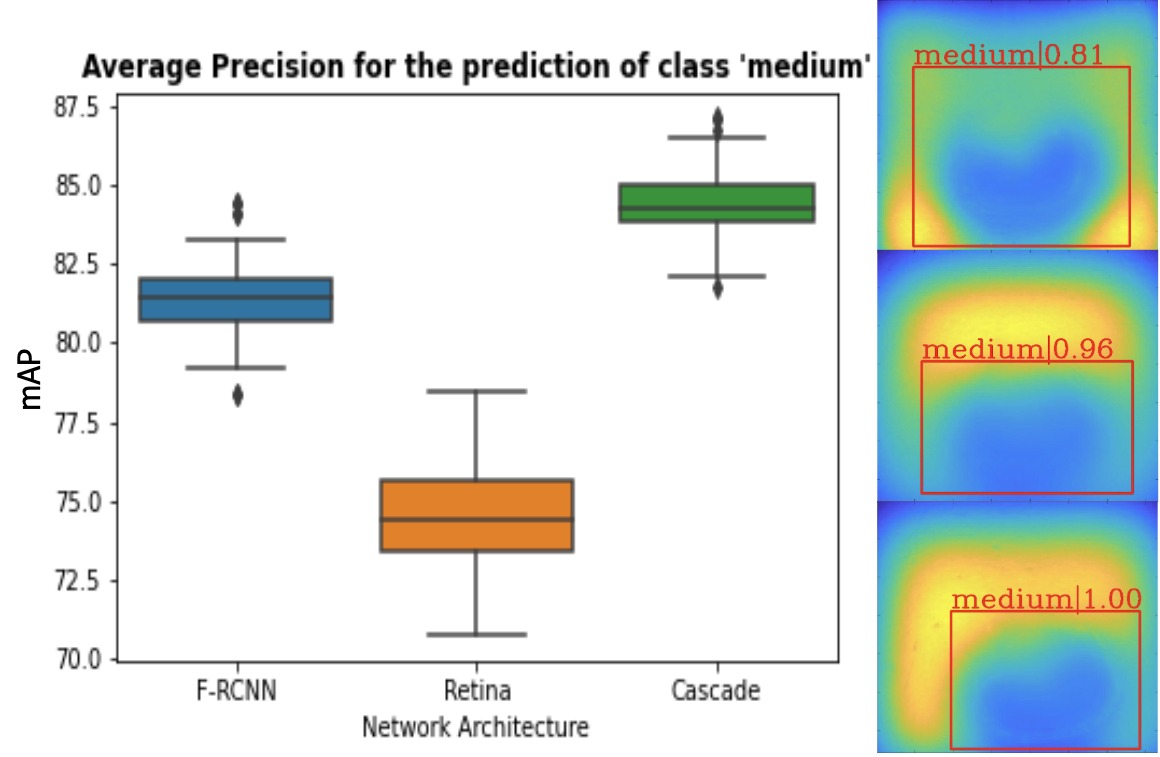}
	\caption{Detection results for class 'medium'}
	\label{res2}
\end{figure}

\begin{figure}[!h]
	\centering
	\includegraphics[width=3.4in]{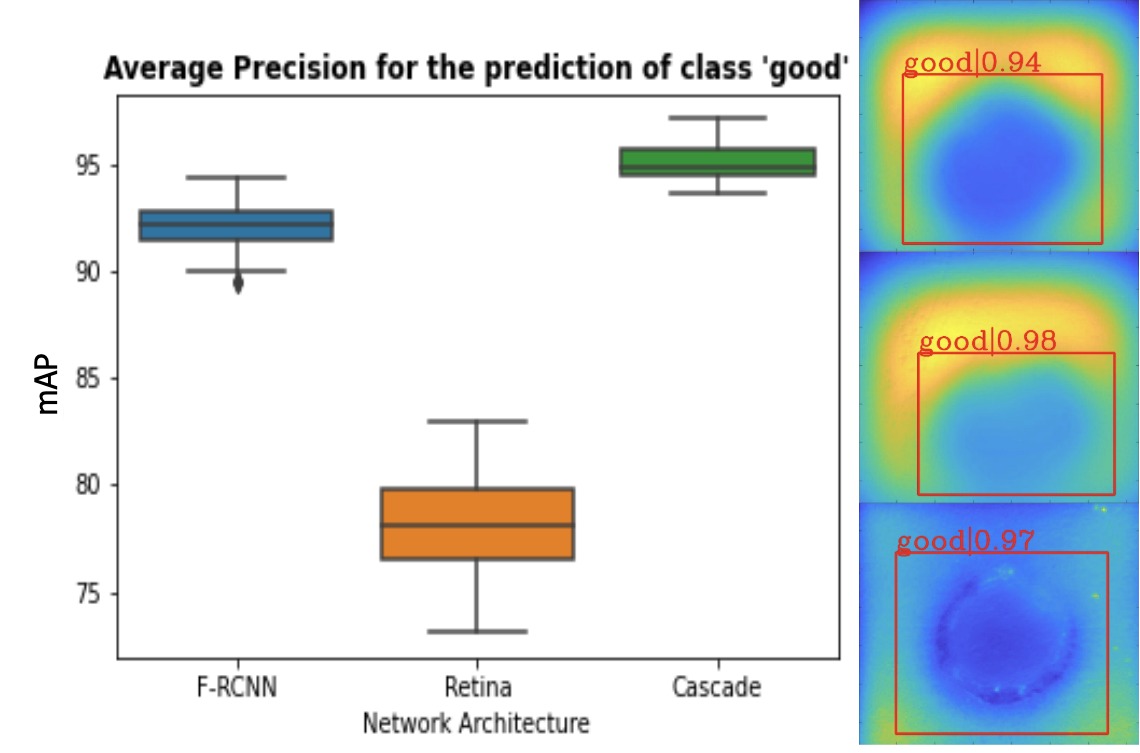}
	\caption{Detection results for class 'bad'}
	\label{res3}
\end{figure}

The dataset generated when applying $F_{10}$ has resulted in robust performance. Based on this, we evaluated two additional network architectures, namely Retina Net and Cascade-RCNN. Furthermore, we evaluated the classfiication accuracies for the three different classes \textit{'good'}, \textit{'medium'} and \textit{'bad'}. Retina contains an additional focal loss function \cite{lin2017focal} while Cascade-RCNN employs an additional network as cascade layer \cite{cai2018cascade}. Table \ref{results} lists relevant metrics of our training for all different approaches. Fig. \ref{res1} to \ref{res3} illustrate the predictions.
\begin{table*}[]\centering
\captionsetup{width=14cm}
	\caption{Evaluation metrics for different architectures. Error rate metric indicates what percentage of all predictions were false. AP metric denotes average likelihood of correct predictions.}
	\begin{tabular}{@{}rrrrcrrrcrrr@{}}\toprule& \multicolumn{3}{c}{\textbf{Faster-RCNN}} & \phantom{abc}& \multicolumn{3}{c}{\textbf{RetinaNet}} & \phantom{abc} & \multicolumn{3}{c}{\textbf{Cascade-RCNN}}\\\cmidrule{2-4} \cmidrule{6-8} \cmidrule{10-12}& 
		$good$ & $medium$ & $bad$ && $good$ & $medium$ & $bad$ && $good$ & $medium$ & $bad$\\\midrule 
		$Error Rate$ &6,33&  36,78&    8,85 &&9,97& 45,6& 12,66&& \textbf{2,72}& 34,66& 16,54 \\
		$mAP$ & 92,14 & 81,57 & 94,234 && 78,12 & 74,21 & 80,97 && \textbf{95,31} & 84,4 & 94,7
		\\\bottomrule
	\end{tabular}
	
	\label{results}
\end{table*}
The error rate metric indicates how many of the predictions were correct and wrong, respectively, with over 90 percent precision. For the average precision metric, we averaged the values of all correct predictions for the different classes. Each bounding box gives a likelihood of the class being predicted, e.g. a value of $0.94$ denotes that the probability of the class is 94 percent. Faster-RCNN and Cascade-RCNN achieve the highest average precision. Especially for the classes \textit{'good'} and \textit{'bad'}, over 95 and 94 percent are achieved, respectively. 
The accuracy and error rates are indicating a stable and reliable prediction for all models. Cascade-RCNN is achieving the best results. A 97 percent accuracy for the class \textit{'good'} and 93 percent for the class \textit{'bad'} is achieved. As expected, the performance is worse for the prediction of the class \textit{'medium'}. Hence, Faster-RCNN achieves a 63 percent accuracy, while Cascade-RCNN achieves an accuracy of 65 percent. The flawed accuracy is due to several reasons: films for the class \textit{medium} were represented the least with only 17 percent. Thus, the imbalance between the different classes \textit{bad} and \textit{good} compared to \textit{medium}, leads to a poor performance. Furthermore, the class \textit{'medium'} is generally hard to visually distinguish from the classes \textit{'good'} and \textit{'bad'}. Remarkably, RetinaNet perform worst in all metrics. This could be attributed to the fact that generally, the welding spots are hard to detect and classify because, even with preprocessing, features are blurry due to the unordered heat distribution throughout the whole image. This makes it hard for classifiers to spot relevant regions. This is enhanced by the fact, that one-stage-detectors rely on one step rather than incorporating an additional region proposal network. Furthermore, large objects are known to cause difficulties for one stage detectors. In our case, the object compromises almost 80 percent of the whole image, which might be another reason why RetinaNet performed worse.

\section{CONCLUSION}
Classifying the quality of spot weldings is a tedious process in industries due to the lack of reliable and robust, non-destructive inspection methods. Common approaches analyze weldings using hand engineered features. Neural networks bear the potential to automate the process and learn relevant features to assess the quality. In this work, we have explored the effect of thermal dataset preparation to generate feasible training datasets for CNNs. Therefore, we take underlying theoretical physical foundations into account and analyzed the intensity value of spot welded joints after pulsed laser thermography. Based on these observations, we proposed data filters and explored their effect on the performance of the CNN. Overall, we could achieve an accuracy of 95 percent in classifying the quality of welds, which motivates not to apply destructive testing methods. Our approach utilizes data generated with laser thermography, which is a cheaper alternative and can be easily applied in-situ, contrary to X-Ray approaches. Additionally, it can be applied for a non-contact inspection, opposing to conventional ultrasonic approaches.
We demonstrated an enhancement by 6 percent when applying our defined data filters, which are based on the maximum intensity area of the film. An important aspect is that smaller data chunks are not sufficient, even with data augmentation, to deliver robust results, and a dataset covering multiple frames is always to be preferred. We also demonstrate the efficiency of different augmentation methods on different areas along the intensity curve. Color augmentation is especially useful for the cooling stage, when the data is similar, while positional augmentation like rotation and flipping can boost accuracy at the earlier stages. Further steps include the modification and optimization of the used neural network models with physics-based optimizers to detect more complex anomalies especially for thermal images. Additionally, we aspire to employ the detection in the frequency domain, which potentially could deliver enhanced results in terms of computational performance and accuracy.



\addtolength{\textheight}{-10cm} 





\bibliographystyle{IEEEtran}

\bibliography{CNNJournal}

\end{document}